# A Histogram Thresholding Improvement to Mask R-CNN for Scalable Segmentation of New and Old Rural Buildings

Ying Li[1], Weipan Xu[1], Haohui "Caron" Chen[2], Junhao Jiang[1], Xun Li[1*]

**Affiliations**

[1]Department of Urban and Regional Planning, School of Geography and Planning, Sun Yat-sen University, Guangzhou, China

[2]Data61, Commonwealth Scientific and Industrial Research Organisation (CSIRO), Australia

*Correspondence to: Xun Li (lixun@mail.sysu.edu.cn)

## Abstract

Mapping new and old buildings are of great significance for understanding socio-economic development in rural areas. In recent years, deep neural networks have achieved remarkable building segmentation results in high-resolution remote sensing images. However, the scarce training data and the varying geographical environments have posed challenges for scalable building segmentation. This study proposes a novel framework based on Mask R-CNN, named HTMask R-CNN, to extract new and old rural buildings even when the label is scarce. The framework adopts the result of single-object instance segmentation from the orthodox Mask R-CNN. Further, it classifies the rural buildings into new and old ones based on a dynamic grayscale threshold inferred from the result of a two-object instance segmentation task where training data is scarce. We found that the framework can extract more buildings and achieve a much higher mean Average Precision (mAP) than the orthodox Mask R-CNN model. We tested the novel framework's performance with increasing training data and found that it converged even when the training samples were limited. This framework's main contribution is to allow scalable segmentation by using significantly fewer training samples than traditional machine learning practices. That makes mapping China's new and old rural buildings viable.

## Introduction

Monitoring the composition of new and old buildings in the rural area is of great significance to rural development [1]. In particular, China's recent rapid urbanization has tremendously transformed its rural settlements over the last decades[2]. However, unplanned and poorly-documented dwellings have posed significant challenges for understanding the rural settlements [3,4]. Traditionally, field surveys had been the major solutions, but they require intensive labour inputs and could be time-consuming, especially in remote areas. The recent breakthroughs of the remote sensing technologies provide growing availability of high-resolution remote sensing images such as low-altitude aerial photos and Unmanned Aerial Vehicle (UAV) images. That allows manual mappings of the rural settlements at a lower cost and with broader coverage, but they are still time-consuming. Therefore, to map the settlements for nearly 564 million rural population in China[5], a scalable, intelligent and image-based solution is urgently needed.

Remote sensing-based mapping of buildings has been a popular research topic for long [6,7,8,9,10]. The pixel-based methods, grouping pixels of similar spectral properties into a particular class, had been widely used in the eras when the remote sensors could only generate images in which pixels are bigger than ground features [11]. Since the launch of IKONO, QuickBird, WorldView and most recently UAVs, the remote sensing images' spatial resolution grows significantly. The pixel-based methods, which do not consider neighboring pixels that are part of the same land cover, failed to utilize different land covers' spatial variation in those high-resolution images. Consequently, the Object-based image analysis (OBIA) has emerged as an effective alternative solution. OBIA use image objects as the basic analysis unit instead of individual pixels [12]. It groups neighboring pixels into shapes with a meaningful representation of the target objects, considering the spatial and hierarchical relationships during the classification process[13]. OBIA involves two main phases: image segmentation and feature classification. The segmentation process divides an image into homogeneous regions, e.g., buildings, water bodies and grasslands, so its quality is critical not just determining the downstream feature classification process but also for the overall performance of the OBIA[14,15]. The recent breakthroughs of machine learning (ML) technologies in computer vision have pushed image segmentation forefronts. ML-based methods, such as Markov Random Fields[16], Bayesian Network[17], Neural Network[18], SVM[19] and Deep Convolution neural network (DCNN) [20] achieved impressive results. These supervised methods learn the spatial and hierarchical relationships between pixels and objects, utilizing remote sensing technologies' resolution gains in recent years. There are two kinds of image segmentation. The semantics segmentation treats multiple objects of the same class as a single entity, while instance segmentation treats multiple objects of the same class as distinct individual entities (or instances) [21,22]. Amongst, Mask R-CNN has been one of the most popular image segmentation methods. It has been used for vehicle-damage-detection , ships labeling, buildings extraction[23,24,25]. The understanding of rural settlements would involve instance segmentation instead of semantic segmentation, as settlement's numbers and areas are needed.

Compared to the urban buildings, rural ones attracted much less attention from the remote sensing community [26,27]. Only a few rural datasets were open to the public. The Wuhan dataset (WHU)[28] extracts 220,000 independent buildings from high-resolution remote sensing images covering 450 km$^2$ in Christchurch, New Zealand. The Massachusetts dataset consists of 151 aerial images of the Boston area, covering roughly 340 km$^2$ [29]. Training on these datasets, the ML-based models achieved impressive segmentation results. However, these data only cover a relatively small area. As a result, the models might not generalize well in other regions where the geographical environments differ significantly. Therefore, the lack of training data specifically for rural environments and the varying geographical environments across regions have posed challenges for building a robust and genialized algorithm. To achieve human-like segmentation for China's vast and varying geography, we might need to build models dedicated to different regions. In this regard, the bottleneck is the manual effort for annotating a large amount of training data. In this study, we proposed a novel framework that could significantly reduce data annotation efforts while retaining the classification capability.

Humans annotate the new and old buildings in high-resolution remote sensing images by the difference of pixel color. This principle can also be applied in the ML algorithm, as the histogram of grayscales would vary significantly across new and old buildings. When new and old buildings' grayscale histogram exhibits bimodal distributions, the valley point can be used as the threshold for discriminating them. This methodology is called histogram thresholding, widely used before the ML prevails[30,31]. If all building footprints are given, a few predicted labels of new and old buildings in the same remote sensing image could to validate if such a bimodal distribution exists and consequently find the valley point. In this regard, we can reduce the number of training samples while retaining the algorithm's capability. It should be noticed that the segmentation of buildings is more manageable than the segmentation of new and old ones in machine learning practices. That is partially due to the reduced efforts for labeling one class instead of two. Another reason is that binary classifiers are more accurate than multi-class classifiers in general. For example, classifying dogs is easier than classifying dog breeds. Therefore, the proposed framework uses histogram thresholding as an add-on to the state-of-the-art deep learning algorithm to achieve impressive segmentation results. In the method section, we will address the proposed framework in detail. This study uses rural areas in Xinxing County, Guangdong Province, as the study case to test the performance of our proposed framework.

## 2. Study Area and Data

## 2.1 Study Area

To test the proposed framework's performance, we collected data samples from high-resolution satellite images covering rural Xinxing County, Guangdong province, China (see Figure 1). Xinxing is a traditional mountainous agricultural county, with a large agricultural population and a relatively complete landscape, forest, and land city. Moreover, Xinxing is a rural revitalization pilot area and has made much rural development and governance achievements. The extraction of new and old buildings is of great significance for understanding rural development in Xinxing.

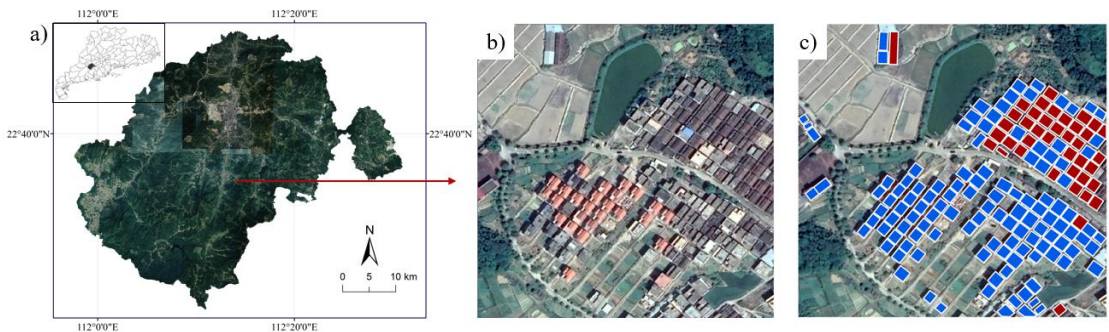

Figure 1 The study area. (a) The location of Xingxing in Guangdong province, China, (b) a village of Xinxing (c) building labels of the same region in panel c (new and old buildings are masked in

## 2.2 Data collection and annotation

Table 1 shows the new and old buildings in high-resolution satellite images. Most of the new buildings are brick-concrete structures, with roofs made of cement or colored tiles. Old houses are mainly Cantonese-style courtyards in Xinxing, where the roof materials are dark tiles. Moreover, the outline of their footprints is less clear than new houses. New buildings are mostly distributed along the streets, while the old buildings still retain a compact comb pattern.

Table 1. the image characteristics of new and old buildings in rural areas

| Type | image characteristics |
|---|---|
| old buildings | 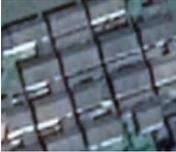 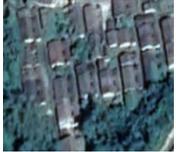 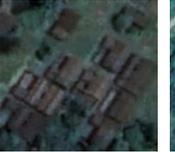 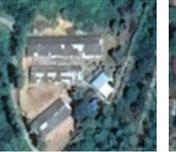 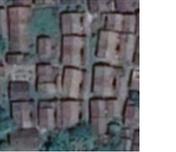 |
| new buildings | 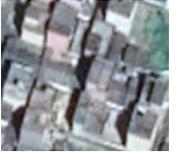 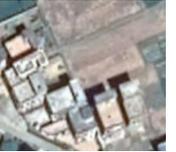 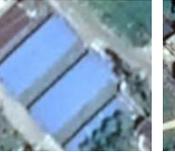 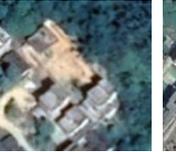 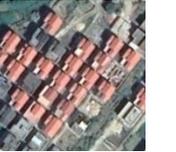 |

We collected 60 images with a resolution of 0.26m. Each image has a size ranging from 900 × 900 to 1024 × 1024 pixels. We use the open-source image annotation tool VIA[32] to delineate the building footprints (see Figure 2). All building samples from those 60 images were compiled as dataset called one-class samples. And We annotated only 26 out of 60 images with new and old labels (called two-class samples hereafter) (see Table 2). Both datasets were spitted into training and validation sets respectively.

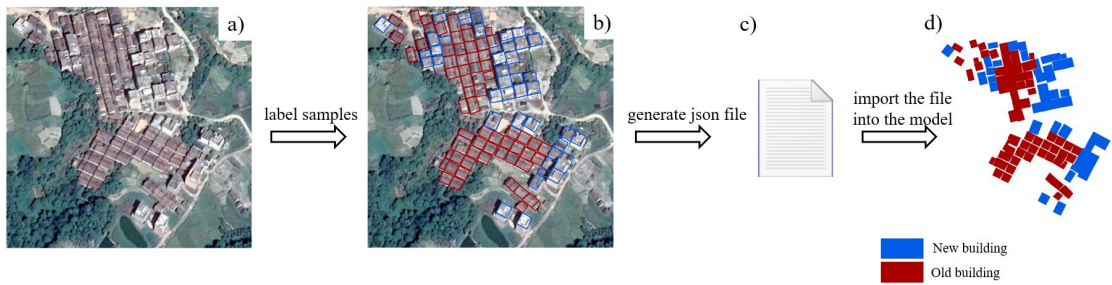

Figure 2. The construction of image dataset. a) a village image; b) the building labels; c) the label json file; d) the mask with building categories.

Table 2. The train and val dataset

|  | one-class samples | two-class samples | | |
|---|---|---|---|---|
|  | buildings | new buildings | old buildings | total |

| | | | | | |
|---|---|---|---|---|---|
| train | 54 pic/5359 poly | 1340 | 1081 | 20 pic/2421 poly | |
| val | 6 pic/892 poly | 498 | 394 | 6 pic/892 poly | |
| total | 60 pic/6251 poly | 1838 | 1475 | 26 pic/3313 poly | |

# 3. Methods

## 3.1 HTMask R-CNN

Mask R-CNN[33] has been proven to be a powerful and adaptable model in many different domains[34,35]. It operates in two phases, generation of region proposals and classification of each generated proposal. In this study, we use Mask R-CNN as our baseline model for benchmarking. As discussed before, we propose a novel segmentation framework that can utilize the histogram thresholding and deep learning's image segmentation capability to extract the new and old rural buildings. We call the proposed framework HTMask R-CNN, abbreviating Histogram Thresholding Mask R-CNN. The workflow of the framework is addressed as follows (see Figure 3 for illustration):

    a. We built two segmentation models (one-class and two-class models) based on the one-class and two-class samples' training sets. The one-class model can extract rural buildings, while the two-class model can classify new and old rural buildings.

    b. An satellite image (Figure 3a) is classified by the one-class and two-class model separately, leading to a map of building footprints (R1 in Figure 3d), and a map of new and old buildings (R2 in Figure 3b).

    c. Grayscale histograms were built using the pixels from the new and old building footprints(R2). The average grayscale levels for new and old buildings were computed as N and O respectively. A valley point is determined by $\theta = (N+O)/2$.

    d. The valley point $\theta$ is used as the threshold to determine the type of building in R1. Finally, we get a map of the old and new buildings in R3(Figure 3e).

The hypothesis is that R3 performs better than R2. Specifically, R3 can take advantage of the capability of R1, while utilizing the grayscale difference of the new and old buildings in R2. The two-class model's performance depends on the numbers of training samples. Assumably, the more the training data are added, the more robust the network training, the better the segmentation results. This study also tests how the numbers of training samples could affect R2 and R3's performance to evaluate how HTMask R-CNN can save the annotation efforts while retaining the segmentation capability.

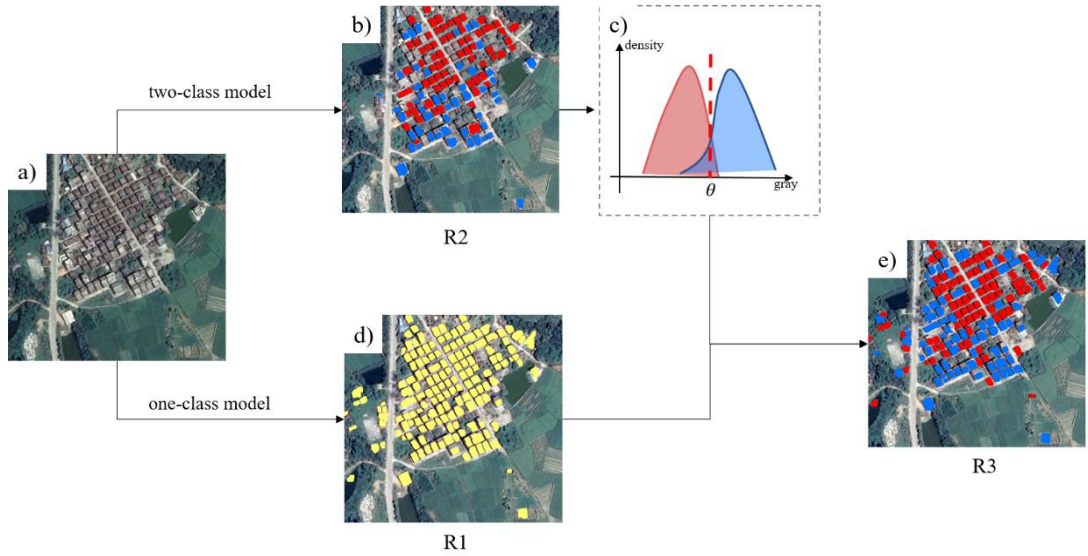

Figure 3. The workflow of HTMask R-CNN. a) a village image; b) R2: the result of two-class model; c) the calculation of exclusive threshold from gray distribution; d) R1:the result of one-class model; e)R3: the result of HTMask R-CNN model.

## 3.2 Experiments

We use R2, the prediction results of the two-class model as the benchmarking. R3 is the result of the proposed framework. We compare R2 and R3 to test how much accuracy improvements in the proposed framework.

We performed data augmentations, including rotating, mirroring, brightness enhancement and adding noise points to the images in both training sets of the one-class and two-class samples. In the training stage for the one-class and two-class models, 50 epochs with two batches per epoch were applied, and the learning rate was set at 0.0001. The Stochastic Gradient Descent (SGD) optimization algorithm was adopted as the optimizer[36]. We set the weight decay to 0.0001, as a penalty added to the loss function to prevent the network model's over-fitting. We set the momentum up to 0.9, which is used to control to what extent the model remains in the original updating direction. We use cross-entropy as a loss function to evaluate the training performance.

To test how HTMask R-CNN can achieve a converged performance with a limited amount of training data, the training process has involved an incremental number of samples (from 5 to 20 satellite images). Afterward, we compare the baseline Mask R-CNN and the HTMask R-CNN by comparing R2 and R3.

## 3.3 Accuracy Assessment

We use the average precision (AP) to quantitatively evaluate our framework on the validation dataset. The AP is equal to taking the area under the precision-recall (PR) curve, equation (1). The mAP$_{50}$ represents the AP value when the threshold of intersection over union (IoU) is 0.5.

$$Precision = TP / (TP + FP)$$
$$Recall = TP / (TP + FN) \quad (1)$$
$$AP = \int_0^1 P(R)dR$$

IoU means the ratio of intersection and union of the prediction and the reference. When a segmentation image is obtained, the value of IoU is calculated according to equation (2).

$$IoU = TP / (TP + FP + FN) \quad (2)$$

## 4. Results

Figure 4 shows the result of an image. In terms of the building footprints mapping, the one-class model has identified most of the buildings. More importantly, it can accurately outline individual buildings and the boundaries of adjacent buildings being correctly separated, which allows the texture of the building to be captured (Figure 4a).

The baseline model (two-class model) performed better and better with the growing numbers of training samples (Figure 4b). However, the numbers of buildings in R2 are still significantly fewer than R1, which aligns with our assumption. In R3, the proposed framework uses R1 as the base map, so the numbers of buildings are equal between R1 and R3. That means the proposed framework outperforms the baseline model.

In terms of the new and old building segmentation, R3 is significantly better than R2 at all levels of training samples. Even when the number of training samples is very limited , e.g., five, the baseline model nearly misidentify between the new and old buildings, while the proposed framework still produce a reasonable result.

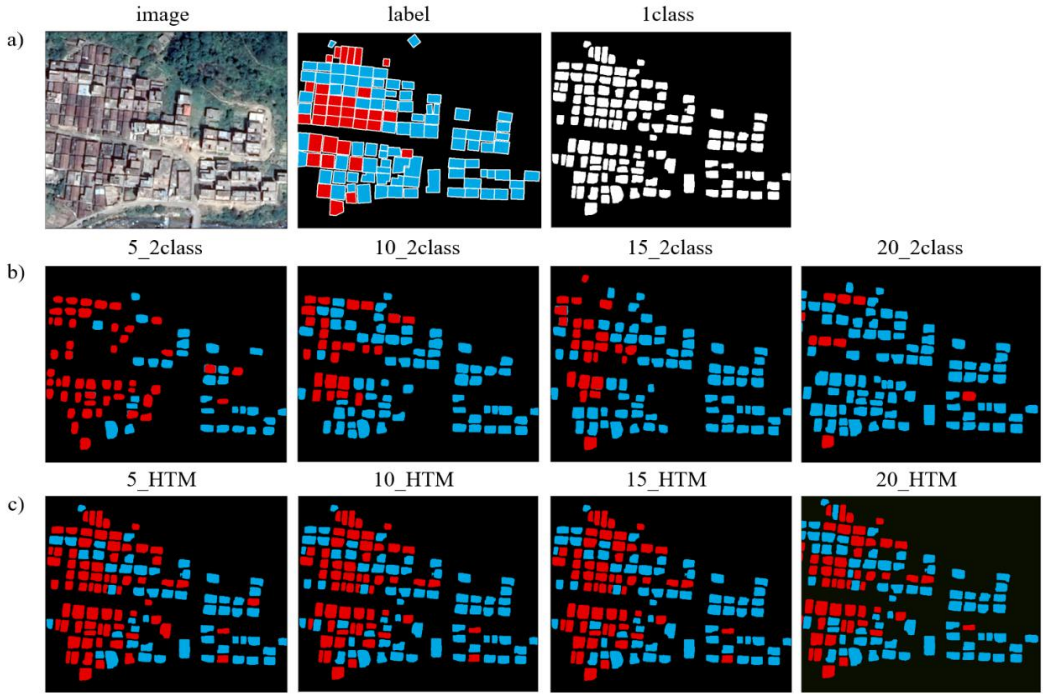

Figure 4. An example shows the result comparison between the baseline Mask R-CNN model and the HTMask R-CNN framework. a) the satellite image from the test set, the second column shows the annotations, and the third column shows the result of one-class model R1. b) the result of the two-class model, R2, with incremental training samples. c) the result of the HTMask R-CNN framework, R3, with incremental training samples.

Figure 5 shows the training process of the one-class model. We noticed that the performance of the one-class model converges at the 25th epoch. Regardless of classification, it identifies most of the buildings, and then the mAP50 could reach 0.70, respectively.

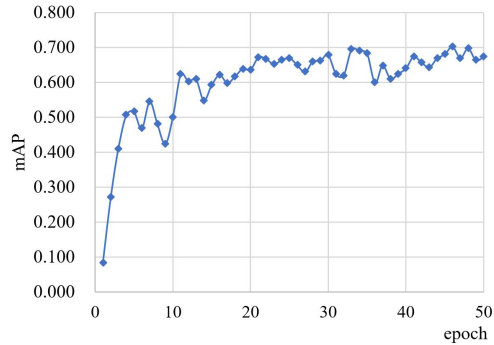

Figure 5 the training process of the one-class model

The baseline two-class model and the HTMask R-CNN also converge at the 25th epoch (Table3, Figure 6). When the training size is small, the mAP50 of the baseline two-class model is very low, while the HTMask R-CNN can significantly improve the recognition. With the increasing training size, the baseline two-class model and HTMask R-CNN's performance gap become narrower(see Figure 5d). Finally, the $mAP_{50}$ of the baseline two-class reached 0.35 when the training size is 20, but still lower than HTMask R-CNN. More importantly, HTMask R-CNN performs consistently (mAP_50≈0.45), no matter which levels of training size.

It confirms our assumption that HTMask R-CNN can perform well in a small number of samples, Which means that it can significantly reduce annotation efforts while retaining the segmentation capability. In contrast , the baseline two-class model performed poorly in extracting old-new two-category buildings.

Table 3. the $mAP_{50}$ between R2 and R3 at all levels of training

| models\epoch | 1 | 5 | 10 | 15 | 20 | 25 | 30 | 35 | 40 | 45 | 50 |
|---|---|---|---|---|---|---|---|---|---|---|---|
| 5_2class | 0.00 | 0.00 | 0.06 | 0.06 | 0.05 | 0.06 | 0.04 | 0.07 | 0.06 | 0.09 | 0.07 |
| 5_HTM | 0.03 | 0.35 | 0.32 | 0.38 | 0.41 | 0.42 | 0.43 | 0.44 | 0.41 | 0.45 | 0.42 |
| 10_2class | 0.00 | 0.01 | 0.09 | 0.12 | 0.15 | 0.17 | 0.16 | 0.17 | 0.20 | 0.22 | 0.22 |
| 10pic_HTM | 0.04 | 0.34 | 0.31 | 0.40 | 0.41 | 0.44 | 0.45 | 0.45 | 0.42 | 0.45 | 0.44 |
| 15_2class | 0.00 | 0.06 | 0.10 | 0.15 | 0.19 | 0.21 | 0.23 | 0.26 | 0.28 | 0.29 | 0.29 |
| 15pic_HTM | 0.04 | 0.34 | 0.33 | 0.39 | 0.42 | 0.44 | 0.45 | 0.48 | 0.43 | 0.45 | 0.45 |
| 20_2class | 0.00 | 0.11 | 0.19 | 0.19 | 0.31 | 0.27 | 0.27 | 0.34 | 0.31 | 0.31 | 0.35 |
| 20_HTM | 0.05 | 0.34 | 0.32 | 0.39 | 0.41 | 0.44 | 0.44 | 0.47 | 0.43 | 0.45 | 0.45 |

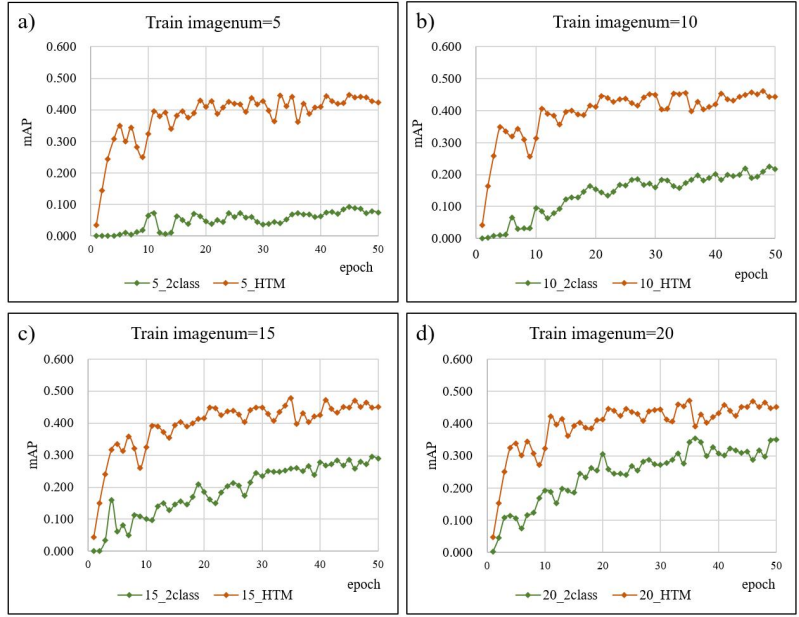

Figure.6 the AP for each models. a). sample size=5; b) sample size=10; c) sample size=15; d) sample size=20.

## 6. Discussions

With the advance of deep learning, the extraction of building footprint from satellite imagery has made notable progress, contributing significantly to settlements' digital records. However, the scarcity of training data has always been the main challenge for scaling building segmentation. Therefore, this study proposes a novel framework based on the Mask R-CNN model and histogram thresholding method to extract old and new rural buildings even when the label is scarce. We tested the framework in Xinxing County, Guangdong province, and achieved promising results. This framework provides a viable solution for mapping China's rural buildings at a significantly reduced cost.

Mask R-CNN models have been proven useful in many applications. However, this study found that the orthodox Mask R-CNN model performed poorly in extracting old-new two-category buildings. When the training samples are limited, the mAP is only 0.35, respectively. We believe the varying geographical environments lead to the poor generalization of the segmentation model when the training samples cannot cover the most distinctive spatial and spectrum features. For instance, the model might not classify a building with an open patio as either a new or old building if none of the training samples contains this unique shape. Meanwhile, the single-category classification task using Mask R-CNN could reach mAP at 0.70, respectively. That means utilizing Mask R-CNN's capability in mapping building footprints could improve the recall rate for the old-new two-category classification task. Hence, we propose such a novel framework.

While tested the framework with increasing training samples, we found that it converges at a very early stage when the numbers of training images are only five. That means the framework can be applied on a large scale, to map all rural buildings in China. Before then, more careful studies should be undertaken to understand the limitations of the framework. For example, more training samples might be needed for an accurate model in other areas.

# 6. Conclusions

Nearly half of the Chinese population live in the rural areas of China. The lack of a digital record of the new and old buildings has posed challenges for the governments to realize the socio-economic state. Under China's central government's current rural revitalization policy, many migrant workers will return to the villages. Therefore, a scalable, intelligent, and accurate building mapping solution is urgently needed. The proposed framework in this study achieved a promising result even when the training samples are scarce. As a result, we can scale the mapping process at a significantly reduced cost. Therefore, we believe this framework could map every settlement in the rural areas, help policymakers establish a longitudinal digital building record, and monitor socio-economics across all rural regions.